\pdfoutput=1

\documentclass{article}

\usepackage{times}
\usepackage{epsfig}
\usepackage{graphicx}
\usepackage{amsmath}
\usepackage{amssymb}

\usepackage{tikz}
\usepackage{url}
\usepackage{rotating}
\usepackage{multirow}
\usepackage{dsfont}
\usepackage{color,colortbl}
\usepackage{pifont}
\usepackage{wrapfig}
\usepackage{algorithm}
\usepackage{flushend}

\usepackage{hyperref}
\PassOptionsToPackage{noend}{algpseudocode}
\usepackage{algpseudocode}
\usepackage{etoolbox}

\usepackage[accepted]{icml2015}

\newcommand{\etal}{\textit{et al}.}
\newcommand{\ie}{\textit{i}.\textit{e}.}
\newcommand{\eg}{\textit{e}.\textit{g}.}

\icmltitlerunning{Hierarchical Maximum-Margin Clustering}

\begin{document} 

\twocolumn[
\icmltitle{Hierarchical Maximum-Margin Clustering}


\icmlauthor{Guang-Tong Zhou}{gza11@cs.sfu.ca}
\icmladdress{Simon Fraser University, Burnaby, Canada}
						
\icmlauthor{Sung Ju Hwang}{sjhwang@unist.ac.kr}
\icmladdress{Ulsan National Institute of Science and Technology, Ulsan, South Korea}

\icmlauthor{Mark Schmidt}{schmidtm@cs.ubc.ca}
\icmladdress{University of British Columbia, Vancouver, Canada}

\icmlauthor{Leonid Sigal}{lsigal@disneyresearch.com}
\icmladdress{Disney Research, Pittsburgh, PA, USA}

\icmlauthor{Greg Mori}{mori@cs.sfu.ca}
\icmladdress{Simon Fraser University, Burnaby, Canada}

\icmlkeywords{hierarchical clustering; maximum-margin learning}

\vskip 0.3in
]

\begin{abstract} 
We present a hierarchical maximum-margin clustering method for unsupervised data analysis.  Our method extends beyond flat maximum-margin clustering, and performs clustering recursively in a top-down manner.  We propose an effective greedy splitting criteria for selecting which cluster to split next, and employ regularizers that enforce feature sharing/competition for capturing data semantics.  Experimental results obtained on four standard datasets show that our method outperforms flat and hierarchical clustering baselines, while forming clean and semantically meaningful cluster hierarchies.
\end{abstract} 

\vspace{-0.15in}

\section{Introduction}
\label{sec:intro}

Clustering is an important topic in machine learning that, after decades of research, remains a challenging and active topic of research.  Clustering aims to group instances together based on their underlying similarity in an unsupervised manner.   Clustering remains an active topic of research due to its widespread applicability in the areas of data analysis, visualization, computer vision, information retrieval, and natural language processing.  Popular clustering methods include k-means clustering~\cite{Lloyd82} and spectral clustering~\cite{Ng01}.

Recent progress in maximum-margin methods has led to the development of maximum-margin clustering (MMC) techniques~\cite{Xu04}, which aim to learn both the separating hyperplanes that separate clusters of data, and the label assignments of instances to the clusters.  MMC outperforms traditional clustering methods in many cases, largely due to the discriminative margin separation criterion imposed among clusters.

However, MMC also has limitations.  First, MMC is not particularly efficient.  While efficient MMC methods have been proposed \cite{Zhang07,Zhao08}, even in such cases the time complexity is at least linear or quadratic with respect to the number of samples and clusters.  This scalability issue is a significant problem when considering the scale of modern datasets.  Second, MMC has difficulty identifying clusters with small margins, which are particularly useful for fine-grained data.  Consider clustering images of commercial vehicles.  In such data the major source of dissimilarity among samples is the viewpoint and this is what MMC is likely to focus on.  The variations in the make of the vehicle, which are semantically more meaningful, would result in more local fine-grained differences and may be ignored by MMC's flat-clustering criterion.

Hierarchical clustering methods, which are typically based on a tree structure, have been extensively studied for their benefits over their flat clustering counterparts.  These hierarchical clustering methods can discover hierarchical structures in data that better represent many real-world data distributions.  Computationally, hierarchical clustering methods are also often more efficient, because one can reduce a single large clustering problem into a set of smaller subproblems to be recursively solved.  Since within each sub-problem the data only needs to be clustered into a small number of clusters, and for lower levels of the hierarchy only a small subset of the data participates in each clustering step, this procedure tends to be a lot more efficient.

To leverage such benefits, we propose a hierarchical extension to MMC that recursively performs k-way clustering in a top-down manner.  However, instead of naively performing MMC at each clustering step, we further leverage the observation from human-defined taxonomies that each grouping/splitting decision typically focuses on different features of the data.


Suppose, again, that we want to cluster different types of commercial vehicles.  Assuming we can cluster the data hierarchically, it is sensible to assume that first we should cluster the data based on the vehicle type (e.g., truck, SUV, sedan).  Once we know which sub-group each instance belongs to, we may want to employ other criteria to separate them, e.g., according to the price range or the make.  We want to leverage a similar intuition to learn clusters that focus on maximizing the margin along different directions at different levels in the hierarchy. Here, directions are defined by subsets of features from the much larger feature vectors describing each instance.  More specifically, we employ regularization that allows clusters to group and compete for the features at different levels.  Such regularization has been made popular in semantic supervised learning in recent years~\cite{Xiao11,Hwang11}, but here we apply the idea in an unsupervised hierarchical clustering framework. 

We test our hierarchical maximum-margin clustering (HMMC) method on several image datasets, and show that HMMC is able to outperform flat clustering methods like MMC.  More significantly, it is able to discover clean and semantically meaningful cluster hierarchies, outperforming other hierarchical clustering alternatives.

Our contributions are threefold: (i) we present a novel hierarchical clustering algorithm based on maximum-margin clustering with an effective greedy splitting criterion for selecting which cluster to split next, (ii) we employ regularization that enforces feature sharing/competition to learn clusters that can focus on important features during clustering, and (iii) we empirically validate that our HMMC can learn semantically meaningful clusters without any human supervision.



\section{Related Work}
\label{sec:related}

\textbf{Maximum-margin clustering}:  MMC was first proposed by Xu et al.~\cite{Xu04}.  It is a maximum-margin method for clustering, analogous to support vector machines (SVMs) for supervised learning problems, that learns both the maximum-margin hyperplane for each cluster and the clustering assignment of instances to clusters.  Since this joint learning results in a non-convex formulation, unlike SVMs, it is often solved by a semidefinite relaxation~\cite{Xu04,Valizadegan06} or alternating optimization~\cite{Zhang07}.  While most of the MMC methods focus on efficient optimization of the non-convex problems, the MMC formulation was also extended to handle the case of multi-cluster clustering problems~\cite{Xu05,Zhao08}, and to include latent variables~\cite{Zhou13}.

\textbf{Hierarchical clustering methods}:  Most hierarchical clustering methods employ either top-down clustering strategies that recursively split clusters into fine-grained clusters, or bottom-up clustering strategies that recursively group the smaller clusters into larger ones~\cite{Manning08}.  Our method is a top-down clustering method, and the canonical example of such a method is hierarchical k-means clustering, which performs k-means recursively in a top-down manner (e.g., the bisecting k-means method~\cite{Steinbach00}).  Variations on this idea include hierarchical spectral clustering (e.g., PDDP~\cite{Boley98}) which performs the hierarchical clustering on the graph Laplacian of the similarity matrix, and model-based hierarchical clustering~\cite{Vaithyanathan00,Castro04,Goldberger04} which fits probabilistic models at each split.  To the best of our knowledge, this is the first work using a maximum-margin approach for hierarchical clustering.

\textbf{Sharing/competing for features}:  Regularization methods that promote certain structures in the parameter or feature spaces have been extensively studied in the context of regression, classification, and sparse coding. The \emph{group} lasso~\cite{Meier08} employs a mixed $\ell_{1,2}$-norm to promote sparsity among groups of features, identifying the groups that are most important for the task.  This has been applied to classification tasks like multi-task learning and multi-class classification, where it encourages the classifier(s) to share features across the tasks/classes. A generalization of the group lasso is the \emph{sparse} group lasso~\cite{Friedman10}, that further encourages sparsity within each individual model.

However, in some cases it makes more sense to have models fit to exclusive sets of features. The \emph{exclusive} lasso~\cite{Zhou10} encourages two models to use different features, by minimizing the $\ell_{2}$-norm of their $\ell_{1}$-norms. This discourages different models from having non-zero values along the same feature dimensions, encouraging each model to use features that are exclusive to their tasks. Orthogonal transfer~\cite{Xiao11} focuses on such exclusiveness between parent and child models in a taxonomy, and enforces the exclusivity through ``orthogonal regularization'' where we minimize the inner product of the SVM weights for parent and child nodes. The tree of metrics approach~\cite{Hwang11} employs similar intuition, but learns Mahalanobis metrics instead of SVM weights, and focuses on selecting sparse and disjoint features.  Tree-guided group lasso~\cite{Kim10} employ both sharing and exclusive regularizations, to promote sharing between the labels that belong to the same parent, while also enforcing exclusive fitting between them, guided by a predefined taxonomy.  These methods consider supervised learning scenarios, while our method utilizes the grouping and exclusive regularizers for unsupervised clustering.

\section{Hierarchical Maximum-Margin Clustering}
\label{sec:method}

We propose a hierarchical clustering method based on the maximum-margin criterion.  We aim to find groups of data points with a large separation between them, while forming a cluster hierarchy.  The proposed method builds on the standard flat MMC clustering~\cite{Xu04}, but extends MMC in the following two aspects: (i) we introduce regularizers to encourage the different layers of the hierarchy to focus on the use of different feature subsets, and (ii) we build the hierarchy iteratively from coarse clusters to fine-grained clusters (rather than forming all clusters in one split) using a greedy top-down algorithm with a novel splitting criterion.  We first introduce the HMMC formulation in this section, and then describe the optimization method in Sec.~\ref{sec:opt}.

Suppose there are $T$ non-leaf nodes $\{n_{t}\}_{t=1}^{T}$ in the learned hierarchy.  We use $\mathcal{D}_{t}$ to denote the data on $n_{t}$, and HMMC splits $\mathcal{D}_{t}$ into $K_{t}$ clusters by learning a linear model $\mathbf{w}_{tk}$ for each cluster $k$.  We collect the $K_{t}$ cluster models in $\mathbf{w}_{t} = \{\mathbf{w}_{tk}\}_{k=1}^{K_{t}}$.  We split the data $\mathcal{D}_{t}$ on node $n_{t}$ using the MMC idea -- finding a clustering assignment such that the resultant margin between clusters is maximal over all possible assignments.  By summing over all the non-leaf splits, our global HMMC objective is formulated as:
\begin{eqnarray}
\label{eqn:hmmc}
\hspace{-0.1in} \min_{\substack{\mathbf{w},\mathbf{y} \\ \xi \geq 0}} & \hspace{-0.1in} & \sum_{t=1}^{T} \Big(\alpha G(\mathbf{w}_{t}) + \beta E(\mathbf{w}_{t}) + \frac{1}{|\mathcal{D}_{t}|K_{t}} \sum_{\substack{\mathbf{x}_{i} \in \mathcal{D}_{t} \\ y \neq y_{ti}}} \xi_{tiy}^{2}\Big), \\
\hspace{-0.1in} s.t. & \hspace{-0.1in} & \mathbf{w}_{ty_{ti}}^{\top}\mathbf{x}_{i} - \mathbf{w}_{ty}^{\top}\mathbf{x}_{i} \geq 1 - \xi_{tiy}, \ \ \ \forall t, \mathbf{x}_{i} \in \mathcal{D}_{t}, y \neq y_{ti} \nonumber\\
\hspace{-0.1in} & \hspace{-0.1in} & y_{ti} \in \{1,\ldots,K_{t}\}, \quad \quad \quad \quad \ \ \ \ \forall t, \mathbf{x}_{i} \in \mathcal{D}_{t} \nonumber\\
\hspace{-0.1in} & \hspace{-0.1in} & L_{t} \leq \sum_{\mathbf{x}_{i} \in \mathcal{D}_{t}} \Delta(y_{ti}=y) \leq U_{t}, \ \ \forall t,y \in \{1,\ldots,K_{t}\} \nonumber
\end{eqnarray}
where $\mathbf{w} = \{\mathbf{w}_{t}\}$ are the cluster model parameters, $y_{ti}$ denotes the cluster label of an instance $\mathbf{x}_{i}$ on node $n_{t}$, $\xi$'s are slack variables to allow margin violations, $G(\cdot)$ and $E(\cdot)$ are regularizers, and $\alpha$ and $\beta$ are trade-off parameters.  Our algorithm uses MMC for each data split, where we enforce the maximum-margin criterion by constraining the score of fitting $\mathbf{x}_{i}$ to its assigned cluster to be sufficiently larger than to any other cluster, using the squared hinge loss (whose smoothness simplifies the optimization).  The last constraint enforces the clusters to be balanced, to avoid degenerate solutions with empty clusters and infinite margins.  Here $\Delta(\cdot)$ is an indicator function, while $L_{t}$ and $U_{t}$ are the lower and upper bounds controlling the size of the clusters.  As suggested in~\cite{Zhou13}, we set $L_{t}$ and $U_{t}$ to $0.9\frac{|\mathcal{D}_{t}|}{K_{t}}$ and $1.1\frac{|\mathcal{D}_{t}|}{K_{t}}$, respectively, to achieve roughly balanced clusters at each split.  Note that HMMC jointly optimizes the model parameters $\mathbf{w}$ and clustering assignments $\mathbf{y} = \{ y_{ti} \}$ over all splits.

The two regularizers $G(\mathbf{w}_{t})$ and $E(\mathbf{w}_{t})$ promote learning of a semantically meaningful cluster hierarchy.  These regularizers encourage splitting on a sparse group of features at each node, but encoding a preference towards using different features at different levels of the hierarchy.  While the grouping and competition among features have proved useful for encoding semantic taxonomies in supervised learning problems~\cite{Xiao11,Hwang11}, we apply these ideas for discovering semantically meaningful cluster hierarchies in an entirely unsupervised setting.

\textbf{Group sparsity}: In the hierarchy, we would like different splits to focus on different subsets of features.  Thus, in splitting a non-leaf node, we encourage the clustering process to only use a sparse set of relevant features.  Considering that there are $K_{t}$ cluster models at node $n_{t}$, we enforce group sparsity over different feature dimensions so that the $K_{t}$ models are using the same subset of features.  Formally, we have the following regularizer on the split of $n_{t}$:
\begin{equation}
\label{eqn:hmmc:group}
G(\mathbf{w}_{t}) = \frac{1}{PK_{t}} \sum_{p = 1}^{P} \sqrt{\sum_{k = 1}^{K_{t}} \mathbf{w}_{tk,p}^{2}},
\end{equation}
where $P$ is the feature dimension, and $\mathbf{w}_{tk,p}$ is the $p$-th element in $\mathbf{w}_{tk}$.  This term encodes that if a feature is irrelevant, then it is zero-weighted in all the $K_{t}$ cluster models.

\textbf{Exclusive sparsity}:  We also want the cluster hierarchy to use different subsets of features in different layers, so that we consider different factors when traversing the hierarchy.
In other words, a split is expected to explore features that are different from its ancestors and descendants, and thus the splits compete for features at different layers.  We will denote a node $n_{t}$'s ancestors by $\mathcal{A}_{t}$, which formally is the set of nodes on the path from the root to $n_{t}$.  With this notation the \emph{exclusive} regularizer for node $n_{t}$ is defined by:
\begin{equation}
\label{eqn:hmmc:exclusive}
E(\mathbf{w}_{t}) = \frac{1}{K_{t}|\mathcal{A}_{t}|P} \sum_{k = 1}^{K_{t}} \sum_{n_{a} \in \mathcal{A}_{t}} \sum_{p=1}^{P} |\mathbf{w}_{tk,p}| \cdot |\mathbf{w}_{ak_{a},p}|,
\end{equation}
where $k_{a}$ indexes the child of $n_{a}$ ($n_{a} \in \mathcal{A}_{t}$) on the path to $n_{t}$.  Thus, $\mathbf{w}_{ak_{a}}$ is the parameter vector for the ancestral cluster to which $n_{t}$ belongs.  Eq.~\eqref{eqn:hmmc:exclusive} penalizes \emph{cooperation} (using the same features) and encourages \emph{competition} (using different features) between a cluster model $\mathbf{w}_{tk}$ and each of its ancestor models $\{\mathbf{w}_{ak_{a}}\}_{n_{a} \in \mathcal{A}_{t}}$.  The degree of competition is calculated as the element-wise multiplication of the absolute weight values.  Intuitively, this means that there is no penalty if two models use different features, but using the same features results in a high penalty.  Consequently, minimizing the exclusive sparsity as we split nodes will encourage nodes to use features different from those used by their ancestors and descendants.  In~\cite{Xiao11,Vervier14}, it is shown that Eq.~\eqref{eqn:hmmc:exclusive} becomes convex when combined with a sufficiently large $\ell_2$-regularizer.


\section{Optimization}
\label{sec:opt}

The objective of Eq.~\eqref{eqn:hmmc} is non-convex due to the unknown hierarchical structure, and because we do not know the split on each node that jointly optimizes $\mathbf{w}$ and $\mathbf{y}$.  To solve the problem, we propose a greedy top-down algorithm to build the hierarchy (Sec.~\ref{sec:opt:hierarchy}), and an alternating descent algorithm for splitting a node (Sec.~\ref{sec:opt:split}).

\subsection{Building the Hierarchy}
\label{sec:opt:hierarchy}

We build the cluster hierarchy in a top-down manner, where the challenge is to iteratively find the next leaf to split.  Algorithm~\ref{alg:greedy} gives an overview of our greedy method.  We start from the root node $n_{1}$ containing all the data.  Note that $n_{1}$ starts as a leaf node since it has no children.  Each iteration tries to split the data on each leaf node $n_{t}$ (Step~\ref{alg:greedy:clustering}), and we define the splitting score (Step~\ref{alg:greedy:splitscore}) as:
\begin{equation}
\label{eqn:hmmc:score}
S(n_{t}) = \frac{\sum_{\mathbf{x}_{i} \in \mathcal{D}_{t}} \mathbf{w}_{ty_{i}}^{\top} \mathbf{x}_{i}}{G(\mathbf{w}_{t}) + E(\mathbf{w}_{t})}.
\end{equation}
The splitting score measures how well, and how easily, the data on node $n_{t}$ can be clustered.  The numerator of Eq.~\eqref{eqn:hmmc:score} summarizes the scores of fitting each instance to its assigned cluster.  A high value in the numerator indicates compact clusters where the instances are well-fit by the assigned cluster models.  The denominator of Eq.~\eqref{eqn:hmmc:score} is the regularization term indicating the complexity of the cluster models, where a small value implies a simple model.  Thus, a higher splitting score means the node is a better candidate to be split.

\begin{algorithm}[t]
\footnotesize
\newcommand{\argmax}{\arg\!\max}

\caption{\small{HMMC: A greedy algorithm for building hierarchy}}
\label{alg:greedy}
\textbf{Input}: \label{alg:greedy:input} $n_{1}$ and $\mathcal{D}$ \Comment{$n_{1}$ is the root node carrying all data in $\mathcal{D}$}

\textbf{Output}: \label{alg:greedy:output} $\mathcal{H}$ \Comment{the cluster hierarchy including all non-leaf nodes}
\begin{algorithmic}[1]
\State \textbf{Initialize}: \label{alg:greedy:ini} $\mathcal{L} \gets \{n_{1}\}$; \Comment{the current set of leaf nodes}
\While{the stopping criterion is not met}
\For{$n_{t} \in \mathcal{L}$} \label{alg:greedy:forleaf}
	\State \label{alg:greedy:clustering} cluster the data on $n_{t}$; \Comment{cf. Sec.~\ref{sec:opt:split}}
	\State \label{alg:greedy:splitscore} compute the splitting score $S(n_{t})$; \Comment{cf. Eq.~\eqref{eqn:hmmc:score}}
\EndFor
\State \label{alg:greedy:greedysplit} $n_{\ast} \gets \argmax_{n_{t} \in \mathcal{L}}S(n_{t})$; \Comment{greedily find the next split}
\State \label{alg:greedy:movenode} $\mathcal{L} \gets \mathcal{L} \setminus n_{\ast}$; $\mathcal{H} \gets \mathcal{H} \cup n_{\ast}$; \Comment{move $n_{\ast}$ from $\mathcal{L}$ to $\mathcal{H}$}
\For{each cluster in $n_{\ast}$} \label{alg:greedy:forcluster} 
	\State \label{alg:greedy:createleaf} create a leaf node $n_{c}$ carrying the data in that cluster; 
	\State \label{alg:greedy:linkchild} link $n_{c}$ as a child of $n_{\ast}$;
	\State \label{alg:greedy:addleaf} $\mathcal{L} \gets \mathcal{L} \cup n_{c}$; \Comment{add $n_{c}$ to the current set of leaf nodes}
\EndFor
\EndWhile
\end{algorithmic}
\end{algorithm}

The leaf node to split is choosen to greedily maximize the splitting score (Step~\ref{alg:greedy:greedysplit}).  We fix the cluster models on this node, mark it as a non-leaf node, and move it to the hierarchy (Step~\ref{alg:greedy:movenode}).  Moreover, since we are splitting this node, we generate its child nodes according to the clustering result and add the child nodes to the leaf node set for the next iteration (Steps~\ref{alg:greedy:forcluster}~to~\ref{alg:greedy:addleaf}).  We iterate this process until the stopping criterion is satisfied, which could test whether (i) a given number of leaf nodes are found, (ii) whether the sizes of all leaf nodes are sufficiently small, or (iii) whether the hierarchy reaches a height limit.  To speed up this process, we cache the clustering result on each leaf node, so that we do not have to rerun the clustering once the leaf node is selected to grow the hierarchy.

\subsection{Splitting A Node}
\label{sec:opt:split}

The clustering on a given node $n_{t}$ is formulated as:
\begin{equation}
\label{eqn:hmmc:split}
\min_{\substack{\mathbf{w}_{t},\mathbf{y}_{t} \\ \xi \geq 0}} \ \ \alpha G(\mathbf{w}_{t}) + \beta E(\mathbf{w}_{t}) + \frac{1}{|\mathcal{D}_{t}|K_{t}} \sum_{\mathbf{x}_{i} \in \mathcal{D}_{t}} \sum_{y \neq y_{ti}} \xi_{tiy}^{2},
\end{equation}
where we omit the constraints (from Eq.~\eqref{eqn:hmmc}) for brevity.  Note that the cluster models of the ancestors of $n_{t}$ have been fixed in the greedy top-down learning process.  Thus, the exclusive regularizer $E(\mathbf{w}_{t})$ becomes a \emph{weighted} $\ell_{1}$-norm (sparsity-inducing) regularizer on $\mathbf{w}_{t}$, where the weight on each model parameter $\mathbf{w}_{tk,p}$ is set based on the ancestor nodes to $\frac{\sum_{n_{a} \in \mathcal{A}_{t}} |\mathbf{w}_{ak_{a},p}|}{K_{t}|\mathcal{A}_{t}|P}$.  Together with the group sparsity $G(\mathbf{w}_{t})$, this yields a weighted sparse group lasso regularizer, generalizing the sparse group lasso regularizer of Friedman \etal~\cite{Friedman10}.

Eq.~\eqref{eqn:hmmc:split} is still a non-convex problem due to the joint optimization over $\mathbf{w}_{t}$ and $\mathbf{y}_{t}$.  We use an alternating descent algorithm to reach a solution.  In each iteration we fix the model parameters $\mathbf{w}_{t}$ and optimize $\mathbf{y}_{t}$ by solving a clustering assignment problem, and then we update $\mathbf{w}_{t}$ while keeping $\mathbf{y}_{t}$ fixed using a proximal quasi-Newton algorithm~\cite{Lee12,Schmidt10}.  The algorithm stops when the objective converges to a local optimum with respect to these steps.

\textbf{Clustering assignment}:  With $\mathbf{w}_{t}$ fixed, the problem in Eq.~\eqref{eqn:hmmc:split} turns out to be an assignment problem, which minimizes the total cost for labeling all instances while maintaining balanced clusters:
\begin{eqnarray}
\label{eqn:hmmc:assign}
\hspace{-0.1in} \min_{\mathbf{y}_{t}} & \hspace{-0.1in} & \sum_{\mathbf{x}_{i} \in \mathcal{D}_{t}} \Delta(y_{ti}=y) \cdot \overbrace{\sum_{y' \neq y} [1 - \mathbf{w}_{ty}^{\top}\mathbf{x}_{i} + \mathbf{w}_{ty'}^{\top}\mathbf{x}_{i}]_{+}^{2}}^{C_{tiy}}, \\
\hspace{-0.1in} s.t. & \hspace{-0.1in} & y_{ti} \in \{1,\ldots,K_{t}\}, \quad \quad \quad \quad \quad \ \ \forall \mathbf{x}_{i} \in \mathcal{D}_{t} \nonumber\\
\hspace{-0.1in} & \hspace{-0.1in} & L_{t} \leq \sum_{\mathbf{x}_{i} \in \mathcal{D}_{t}} \Delta(y_{ti}=y) \leq U_{t}, \quad \forall y \in \{1,\ldots,K_{t}\} \nonumber
\end{eqnarray}
where $C_{tiy}$ is the cost for assigning an instance $\mathbf{x}_{i}$ into a cluster $y$.  Following~\cite{Zhou13}, we could solve Eq.~\eqref{eqn:hmmc:assign} by constructing an integer linear programming (ILP) problem with $O(|\mathcal{D}_{t}| \cdot K_{t})$ variables and $O(|\mathcal{D}_{t}|+K_{t})$ constraints. However, this ILP is time-consuming since in the worst case the complexity of existing ILP solvers is exponential in the number of variables. To efficiently solve this problem, we formulate it as a minimum cost flow (MCF) problem.

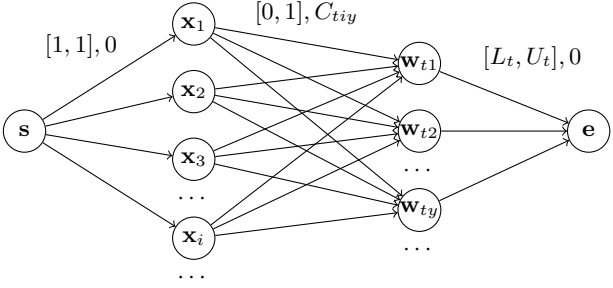
\begin{figure}[t]
\small

\tikzstyle{vertex} = [shape = circle, draw, minimum size = 16pt, inner sep = 0pt]
\tikzstyle{etc vertex} = [shape = circle, minimum size = 5pt, inner sep = 0pt]
\tikzstyle{edge} = [draw,->]
\tikzstyle{weight} = [font=\small]

\centering
\begin{tikzpicture}[scale=1.5, auto] 
\node[vertex] (s) at (-0.5,0.25) {$\mathbf{s}$};
\node[vertex] (e) at (4.5,0.25) {$\mathbf{e}$};

\node[vertex] (x1) at (1,1.2) {$\mathbf{x}_{1}$};
\node[vertex] (x2) at (1,0.6) {$\mathbf{x}_{2}$};
\node[vertex] (x3) at (1,0) {$\mathbf{x}_{3}$};
\node[etc vertex] (xe) at (1,-0.35) {$\cdots$};
\node[vertex] (xn) at (1,-0.7) {$\mathbf{x}_{i}$};
\node[etc vertex] (xee) at (1,-1.05) {$\cdots$};

\node[vertex] (w1) at (3,0.85) {$\mathbf{w}_{t1}$};
\node[vertex] (w2) at (3,0.25) {$\mathbf{w}_{t2}$};
\node[etc vertex] (we) at (3,-0.1) {$\cdots$};
\node[vertex] (wk) at (3,-0.45) {$\mathbf{w}_{ty}$};
\node[etc vertex] (wee) at (3,-0.8) {$\cdots$};

\path[edge] (s) -- node {} (x1);
\path[edge] (s) -- node {} (x2);
\path[edge] (s) -- node {} (x3);
\path[edge] (s) -- node {} (xn);

\path[edge] (x1) -- node {} (w1);
\path[edge] (x1) -- node {} (w2);
\path[edge] (x1) -- node {} (wk);
\path[edge] (x2) -- node {} (w1);
\path[edge] (x2) -- node {} (w2);
\path[edge] (x2) -- node {} (wk);
\path[edge] (x3) -- node {} (w1);
\path[edge] (x3) -- node {} (w2);
\path[edge] (x3) -- node {} (wk);
\path[edge] (xn) -- node {} (w1);
\path[edge] (xn) -- node {} (w2);
\path[edge] (xn) -- node {} (wk);

\path[edge] (w1) -- node {} (e);
\path[edge] (w2) -- node {} (e);
\path[edge] (wk) -- node {} (e);

\node[etc vertex] (cc1) at (0,1) {$[1,1],0$};
\node[etc vertex] (cc2) at (2,1.3) {$[0,1],C_{tiy}$};
\node[etc vertex] (cc3) at (4,0.9) {$[L_{t},U_{t}],0$};
\end{tikzpicture}

\caption{A sample MCF network. The edge settings are formatted as: ``[\emph{lower capacity}, \emph{upper capacity}], \emph{cost}''.  See text for details.}
\label{fig:mcf}
\end{figure}

We re-write the clustering assignment as the problem of sending an MCF through an appropriately designed network, illustrated in Fig.~\ref{fig:mcf}.  The flow capacity of an edge from the starting node $\mathbf{s}$ to an instance node $\mathbf{x}_{i}$ is set to 1 since we are assigning every instance into a cluster.  This one unit of flow is sent from $\mathbf{x}_{i}$ to a cluster node $\mathbf{w}_{ty}$, to which the instance is assigned, at cost $C_{tiy}$.  Finally, each cluster node sends its receiving flows to the end node $\mathbf{e}$, where we limit the flow capacity in the range $[L_{t},U_{t}]$ to take the cluster balance constraints into account.  It can be shown that clustering the $|\mathcal{D}_{t}|$ instances into $K_{t}$ clusters (under the cluster balance constraints) is equivalent to sending $|\mathcal{D}_{t}|$ units of flow from $\mathbf{s}$ to $\mathbf{e}$, and the optimal network flow corresponds to the minimum total cost of Eq.~\eqref{eqn:hmmc:assign}.  To find this optimal flow, we apply the capacity scaling algorithm~\cite{Edmonds72} implemented in the LEMON library~\cite{Dezs11}, which is an efficient dual solution method running in $O\big(|\mathcal{D}_{t}| \cdot K_{t} \cdot \log(|\mathcal{D}_{t}| + K_{t}) \cdot \log(U_{t} \cdot |\mathcal{D}_{t}| \cdot K_{t})\big)$ complexity.  In practice, our MCF solver speeds up the ILP solver in~\cite{Zhou13} by 10 to 100 times.

\def\norm#1{\|#1\|}
\def\b#1{\mathbf{#1}}

\textbf{Updating $\textbf{w}_{t}$}:  With fixed $\mathbf{y}_{t}$, we solve for $\mathbf{w}_{t}$ (a convex problem) using a proximal quasi-Newton method~\cite{Lee12,Schmidt10}. This method is designed to efficiently minimize smooth losses with non-smooth but simple regularizers, and on each iteration it computes a new estimate $\textbf{w}_t$ by solving:
\begin{eqnarray}
\label{eqn:hmmc:pqn}
\hspace{-0.1in} \min_{\textbf{w}_{t}} & \hspace{-0.1in} & \alpha G(\textbf{w}_{t}) + \beta E(\textbf{w}_{t}) + H(\textbf{w}_{t}^{old}) \\
\hspace{-0.1in} & \hspace{-0.1in} & + H'(\textbf{w}_{t}^{old})^\top(\textbf{w}_{t} - \textbf{w}_{t}^{old}) + \frac{1}{2s} \norm{\textbf{w}_{t}-\textbf{w}_{t}^{old}}_{B}^{2}, \nonumber
\end{eqnarray}
where $s$ is a step-size set using a backtracking line-search, $H(\textbf{w}_{t}^{old})$ is the squared hinge-loss (\ie, the last term of Eq.~\eqref{eqn:hmmc:split} after using the constraints to eliminate the slack variables) estimated with $\textbf{w}_{t}^{old}$ from the fixed $\mathbf{y}_{t}$, $H'(\textbf{w}_{t}^{old})$ is the derivative of $H(\textbf{w}_{t}^{old})$ w.r.t. $\textbf{w}_{t}^{old}$, and $\norm{\mathbf{z}}_{B}^{2} = \mathbf{z}^{\top}B\mathbf{z}$ is a divergence formed using the L-BFGS matrix $B$~\cite{Byrd94,Nocedal80}.

A spectral proximal-gradient method is used to compute an approximate minimizer of this objective.  This algorithm requires the proximal operator.  For our weighted sparse group lasso regularizer, we can show that solving this minimizing problem involves a two-step procedure. First, we incorporate the weighted $\ell_{1}$-norm penalty by applying the soft-threshold operator $\mathbf{w}_{tk,p} = \frac{\mathbf{w}_{tk,p}}{|\mathbf{w}_{tk,p}|} [|\mathbf{w}_{tk,p}| - s\beta\lambda_{E}]_{+}$ to each model parameter individually, where $\lambda_{E} = \frac{\sum_{n_{a} \in \mathcal{A}_{t}} |\mathbf{w}_{ak_{a},p}|}{K_{t}|\mathcal{A}_{t}|P}$ is the weights coming from the ancestor models in $E({\mathbf{w}_{t}})$.  This operator returns $0$ if $\mathbf{w}_{tk,p}=0$.  Second, we incorporate the group sparsity using the group-wise soft-threshold operator $\mathbf{w}_{t:,p} = \frac{\mathbf{w}_{t:,p}}{\norm{\mathbf{w}_{t:,p}}_{2}} [\norm{\mathbf{w}_{t:,p}}_{2} - s\alpha\lambda_{G}]_{+}$, where $\mathbf{w}_{t:,p}=[\mathbf{w}_{t1,p},\ldots,\mathbf{w}_{tK_{t},p}]^{\top}$ is the grouping of $K_{t}$ cluster models on a feature dimension $p$, and $\lambda_{G} = \frac{1}{PK_{t}}$ is the normalization term from $G({\mathbf{w}_{t}})$.  Note that this operator returns $\mathbf{0}$ if $\mathbf{w}_{t:,p} = \mathbf{0}$.

\textbf{Convergence analysis}:  We now show that this alternating descent algorithm converges to a local optimum.  The optimization consists of two alternating steps: updating the discrete $\mathbf{y}_{t}$ and the continuous $\mathbf{w}_{t}$.  In the $\mathbf{w}_{t}$ update, we fix the clustering $\mathbf{y}_{t}$ and use a method that is guaranteed to find a global optimum~\cite{Lee12,Schmidt10}.  The $\mathbf{y}_{t}$ update (with $\mathbf{w}_{t}$ fixed) is NP-hard but we can find a solution that guarantees improvement using MCF.  Since there is a finite number of possible assignments to $\mathbf{y}_{t}$, the procedure guarantees convergence to a local minimum with respect to updating $\mathbf{w}_{t}$ or $\mathbf{y}_{t}$.

\section{Experiments}
\label{sec:exper}

\textbf{Datasets}: We evaluate the performance of HMMC on four datasets from two public image collections: Animal With Attributes (AWA)~\cite{Lampert09}  and ImageNet~\cite{Deng09}.  Both collections have natural hierarchies consisting of fine-grained image classes that can be grouped into more general classes.

AWA contains 30,475 images from 50 animal classes (\eg, bat and deer).  We use two datasets following the practice of~\cite{Hwang11}.  The first one, AWA-ATTR, has 85 features consisting of the outputs of 85 linear SVMs trained to predict the presence/absence of the 85 nameable properties annotated by~\cite{Lampert09}, like red and furry.  The second dataset, AWA-PCA, uses the provided features (SIFT, rgSIFT, PHOG, SURF, LSS, RGB) after being concatenated, normalized, and PCA-reduced to 100 dimensions.  The ground-truth hierarchy of AWA is shown in Fig.~2 of~\cite{Hwang11}.

We use two datasets collected from ImageNet: VEHICLE contains 20 vehicle classes (\eg, cab and canoe) and 26,624 images~\cite{Hwang11}, and IMAGENET consists of 28,957 images spanning 20 non-animal, non-vehicle classes (\eg, lamp and drum)~\cite{Hwang12}.  The raw image features are the provided bag-of-words histograms obtained by SIFT~\cite{Deng10, Deng09}.  We also project them down to 100 dimensions with PCA.  The semantic hierarchies of VEHICLE and IMAGENET are given in Fig.~3 of~\cite{Hwang11} and Fig.~2(e) of~\cite{Hwang12}, respectively.


\textbf{Baselines}:  We compare HMMC with four sets of baselines.  The first set is the flat clustering methods k-means (KM), spectral clustering (SC)~\cite{Ng01}, and an MMC approach implemented in~\cite{Zhou13}.

The second set is hierarchical bottom-up clustering (HBUC).  We have tested a variety of methods including Single-Link (SL), Average-Link (AL) and Complete-Link (CL)~\cite{Manning08}.  The pairwise dissimilarity between two images is measured by Euclidean distance.

The third set is hierarchical top-down clustering methods (HTDC).  We derive variants of hierarchical k-means (HKM) and hierarchical spectral clustering (HSC) directly from our HMMC approach.  HKM and HSC apply the same greedy top-down approach as HMMC, but split a given node using k-means and spectral clustering, respectively.  Similar to HMMC, HKM and HSC first try splitting all the current leaf nodes, and then greedily grow the leaf with the best splitting.  The splitting score on a leaf node is defined as the average within-cluster distance -- minimizing this gives the most compact clusters.  We also considered two other baselines, HKM-D and HSC-D.  Instead of growing the leaf with the most compact clusters, HKM-D and HSC-D grow the leaf with the most scattered data, which is defined as the total distance of all instances to their center.

The fourth set of baselines are variants of HMMC.  We change the regularization to derive HMMC-G (group sparsity only), HMMC-E (exclusive sparsity only), HMMC-1 (basic $\ell_{1}$-norm), and HMMC-2 (squared $\ell_{2}$-norm).

\begin{table*}[t]
\scriptsize
\tabcolsep 4.5pt
\centering
\begin{tabular}{|c|c||c|c|c|r||c|c|c|r||c|c|c|r||c|c|c|r|}
\multicolumn{2}{c}{} & \multicolumn{4}{c}{AWA-ATTR} & \multicolumn{4}{c}{AWA-PCA} & \multicolumn{4}{c}{VEHICLE} & \multicolumn{4}{c}{IMAGENET} \\
\hline
\multicolumn{2}{|c||}{Methods} & SP & PS & RI & runtime & SP & PS & RI & runtime & SP & PS & RI & runtime & SP & PS & RI & runtime \\
\hline
\hline
\multirow{3}{*}{\begin{sideways} \makebox[0.2in][c]{FLAT} \end{sideways}}
& KM & 77.95 & \textbf{92.83} & \textbf{96.04} & 5.2 & 77.48 & \textbf{91.34} & 94.50 & 7.4 & 75.44 & 76.76 & 78.08 & 2.9 & 79.66 & 82.03 & 87.14 & 4.0\\
\cline{2-18}
& SC & 77.90 & 92.54 & 95.71 & 209.2 & 77.16 & 88.72 & 91.21 & 172.2 & 74.15 & 74.00 & 74.12 & 112.0 & 69.39 & 67.79 & 61.25 & 137.3\\
\cline{2-18}
& MMC & 77.08 & 83.67 & 83.71 & 15957.7 & 77.32 & 90.59 & 93.54 & 6077.0 & 78.03 & 84.23 & 88.49 & 1366.6 & 79.98 & 82.74 & 89.24 & 2634.3\\
\hline
\hline
\multirow{3}{*}{\begin{sideways} \makebox[0.2in][c]{HBUC} \end{sideways}}
& SL & 63.97 & 16.24 &  2.65 & 88.3 & 64.00 & 16.24 &  2.69 & 80.9 & 58.21 & 29.12 &  5.30 & 61.2 & 45.85 & 32.97 &  5.24 & 84.5\\
\cline{2-18}
& AL & 74.55 & 38.99 & 32.84 & 72.1 & 64.37 & 17.00 &  3.94 & 80.9 & 58.24 & 29.17 &  5.45 & 47.8 & 45.87 & 33.00 &  5.29 & 49.0\\
\cline{2-18}
& CL & 92.60 & 87.54 & 93.33 & 81.8 & 68.14 & 22.63 & 34.27 & 47.6 & 58.30 & 29.26 &  5.58 & 54.2 & 46.46 & 33.62 &  6.59 & 69.3\\
\hline
\hline
\multirow{5}{*}{\begin{sideways} \makebox[0.2in][c]{HTDC} \end{sideways}}
& HKM & 71.95 & 40.46 & 30.02 & 1.6 & 85.00 & 76.86 & 79.77 & 3.2 & 77.68 & 65.75 & 59.75 & 2.2 & 82.85 & 80.93 & 84.67 & 1.9\\ 
\cline{2-18}
& HSC & 81.59 & 69.84 & 67.43 & 247.0 & 79.47 & 47.69 & 57.25 & 873.2 & 68.59 & 45.84 & 36.62 & 745.4 & 64.76 & 52.69 & 53.64 & 913.6\\ 
\cline{2-18}
& HKM-D & 92.59 & 91.01 & 95.97 & 5.8 & 91.43 & 88.24 & 95.02 & 2.4 & 84.89 & 74.77 & 85.37 & 1.9 & 81.42 & 80.87 & 86.60 & 3.2\\ 
\cline{2-18}
& HSC-D & 94.18 & 90.38 & 95.98 & 293.4 & 79.94 & 48.02 & 57.97 & 873.4 & 69.29 & 46.07 & 37.11 & 316.9 & 48.19 & 37.01 & 11.00 & 892.9\\
\cline{2-18}
& HMMC & \textbf{94.40} & 91.03 & 95.96 & 1986.9 & \textbf{93.69} & 89.66 & \textbf{95.65} & 1550.1 & \textbf{90.48} & \textbf{85.08} & \textbf{90.16} & 994.3 & \textbf{86.94} & \textbf{84.63} & \textbf{90.59} & 1411.6\\
\hline
\hline
\multirow{4}{*}{\begin{sideways} \makebox[0.2in][c]{VARIANT} \end{sideways}}
& HMMC-G & 94.36 & 90.83 & 95.87 & 1389.6 & 93.77 & 89.59 & 95.56 & 1254.2 & 90.40 & 85.03 & 90.10 & 883.6 & 86.69 & 84.33 & 90.56 & 1016.4\\
\cline{2-18}
& HMMC-E & 93.81 & 90.74 & 95.45 & 788.2 & 93.11 & 89.39 & 94.79 & 1408.8 & 87.82 & 83.00 & 86.26 & 616.3 & 85.77 & 83.98 & 89.06 & 1658.4\\
\cline{2-18}
& HMMC-1 & 87.77 & 77.49 & 77.84 & 558.1 & 92.01 & 89.69 & 95.49 & 769.9 & 89.94 & 84.72 & 89.54 & 486.9 & 86.81 & 84.52 & 90.52 & 690.1\\
\cline{2-18}
& HMMC-2 & 92.70 & 90.92 & 95.99 & 893.2 & 93.65 & 89.47 & 95.25 & 1449.1 & 90.05 & 84.71 & 89.68 & 541.3 & 86.13 & 84.08 & 89.94 & 1158.3\\
\hline
\end{tabular}
\caption{Clustering performance on the four datasets. SP, PS and RI are reported in percentage, and the boldfaced numbers achieve the best performance among flat and hierarchical methods (excluding HMMC variants).  The runtime (in seconds) is measured on a machine with Intel Xeon 2.8GHz CPU and 16GB memory.}
\label{table:performance}
\end{table*}

\begin{figure*}[t]
\tabcolsep 0pt
\small
\centering
\begin{tabular}{cccc}
\includegraphics[width=0.25\textwidth]{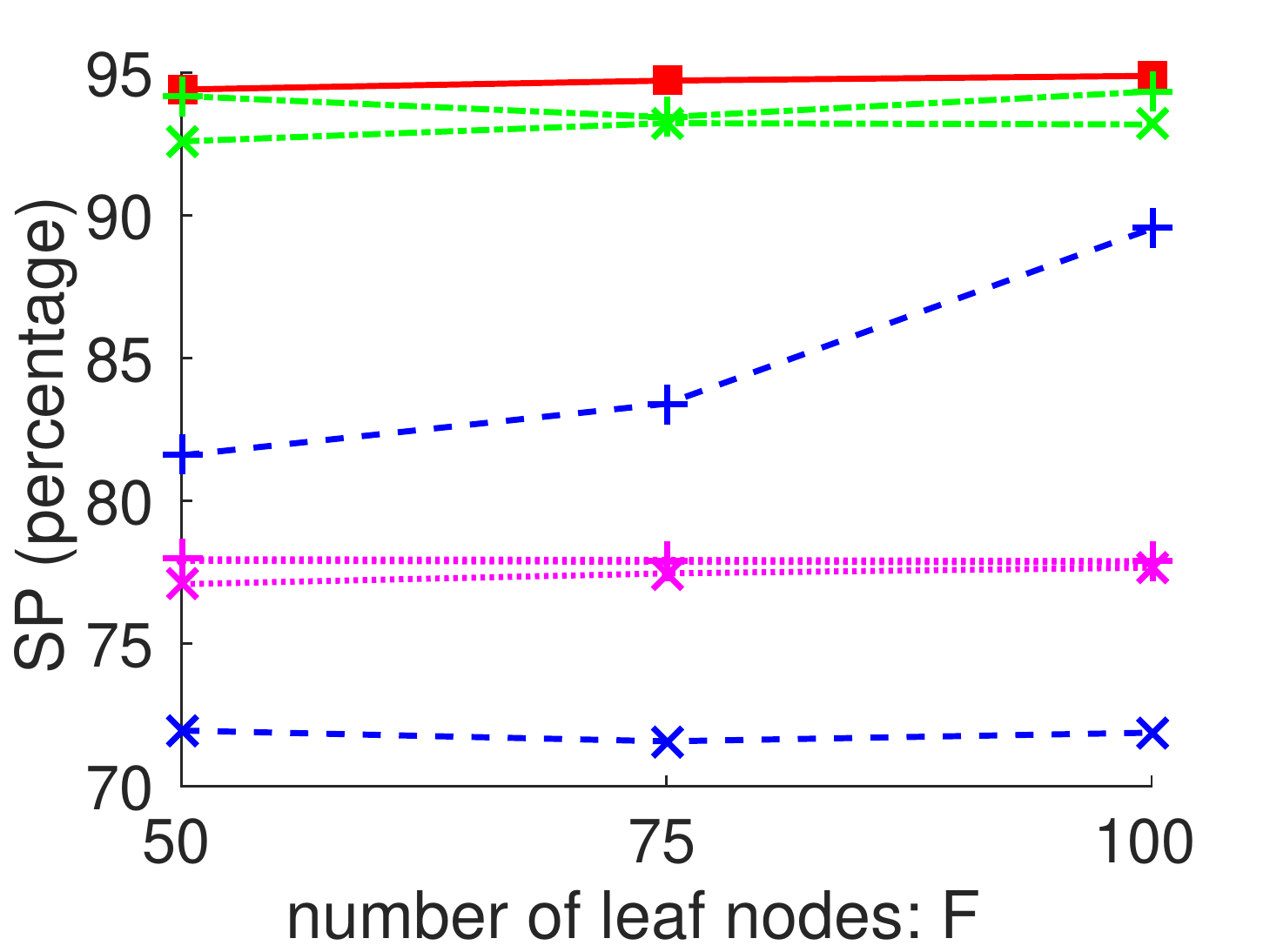} & 
\includegraphics[width=0.25\textwidth]{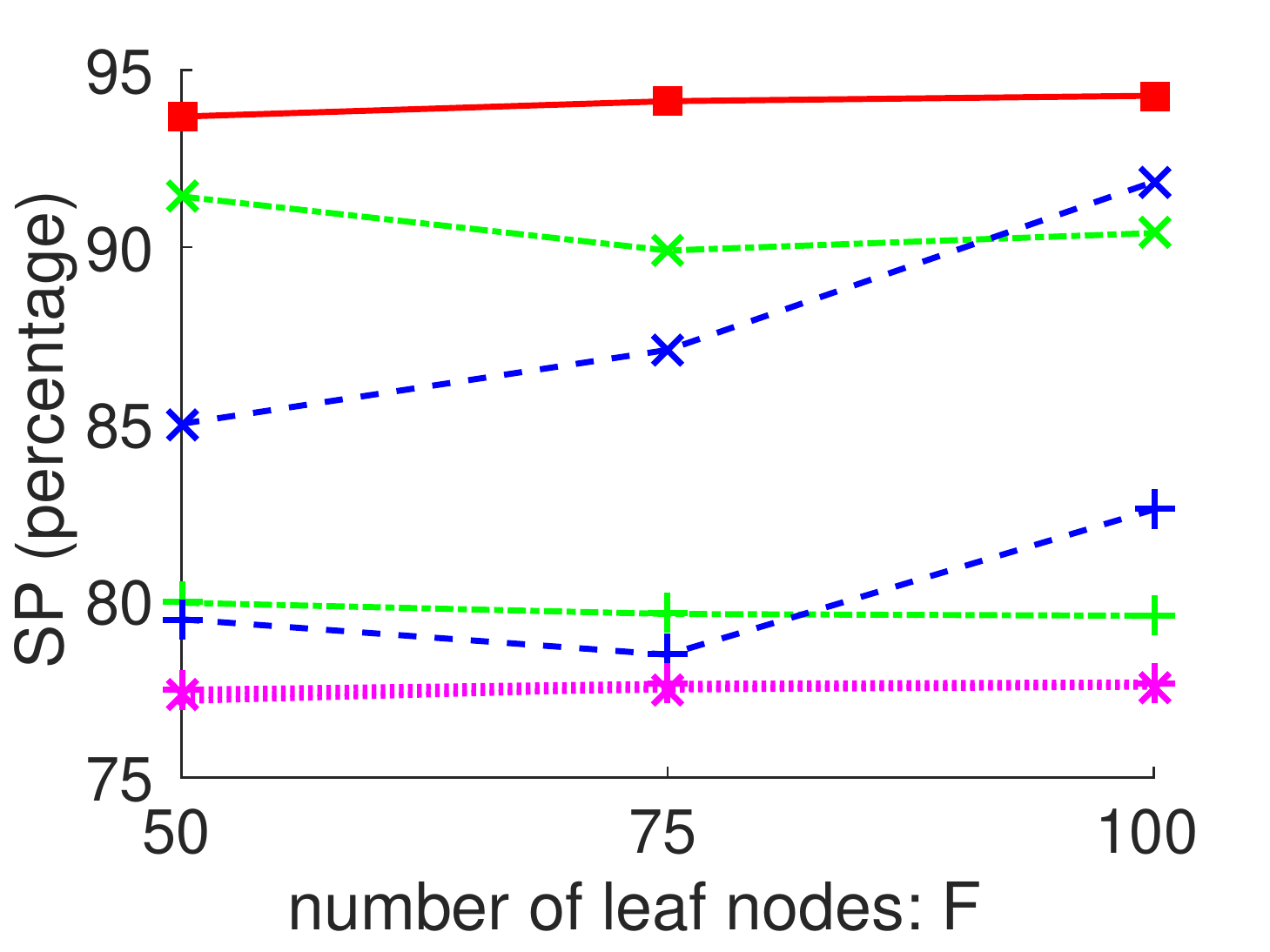} & 
\includegraphics[width=0.25\textwidth]{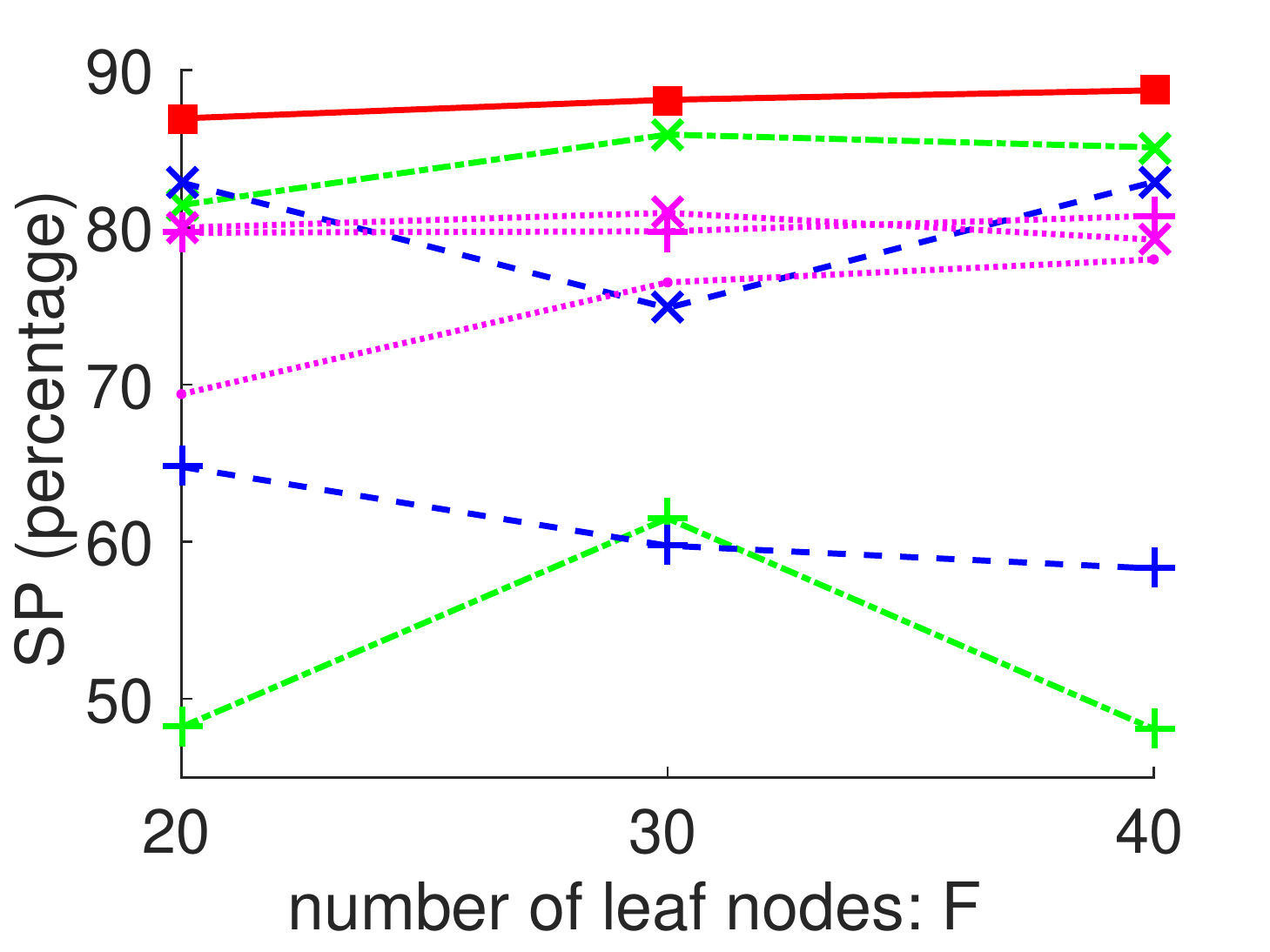} & 
\includegraphics[width=0.25\textwidth]{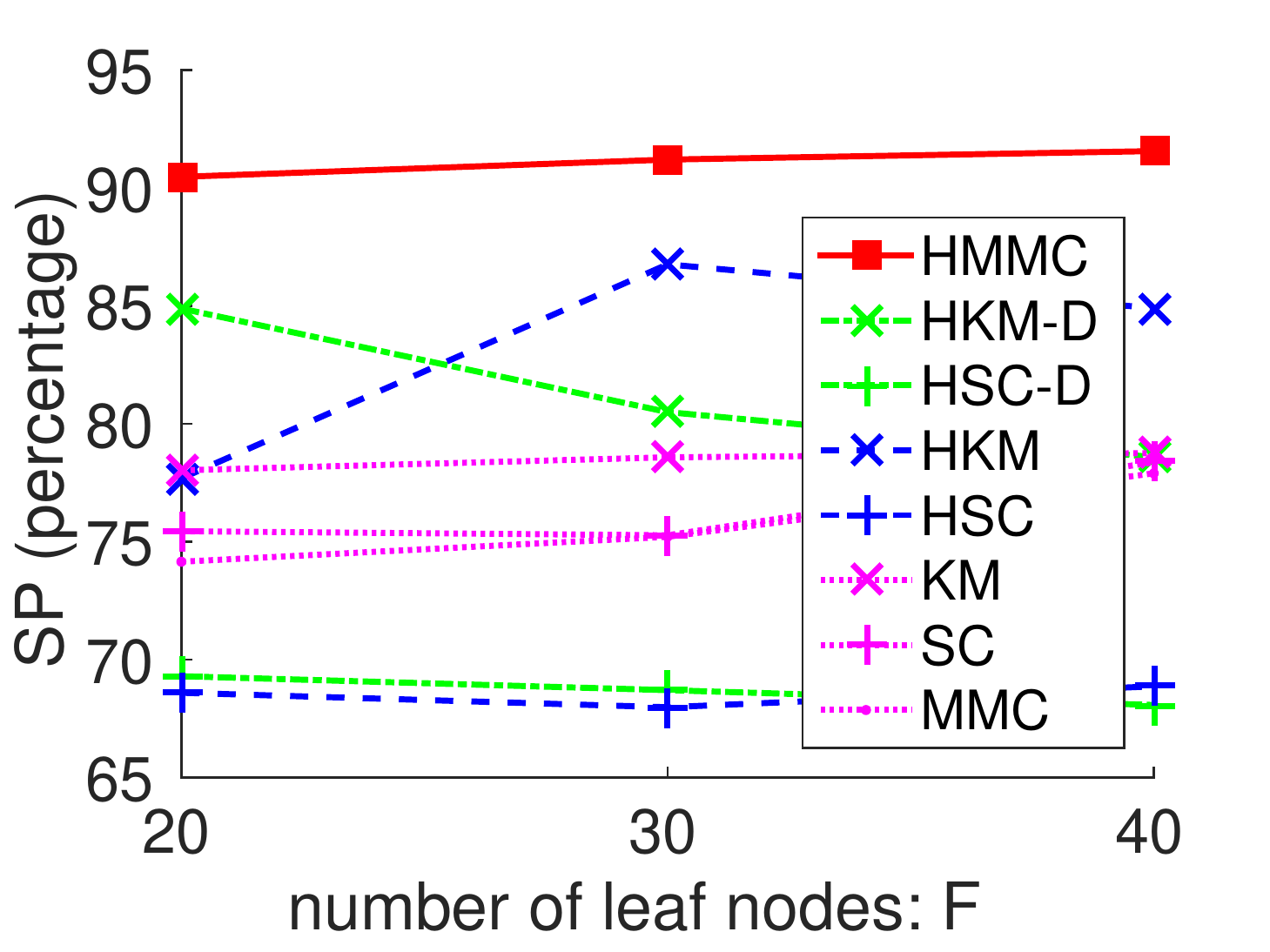} \\
(a) $F$ on AWA-ATTR & (b) $F$ on AWA-PCA & (c) $F$ on VEHICLE & (d) $F$ on IMAGENET \\
\includegraphics[width=0.25\textwidth]{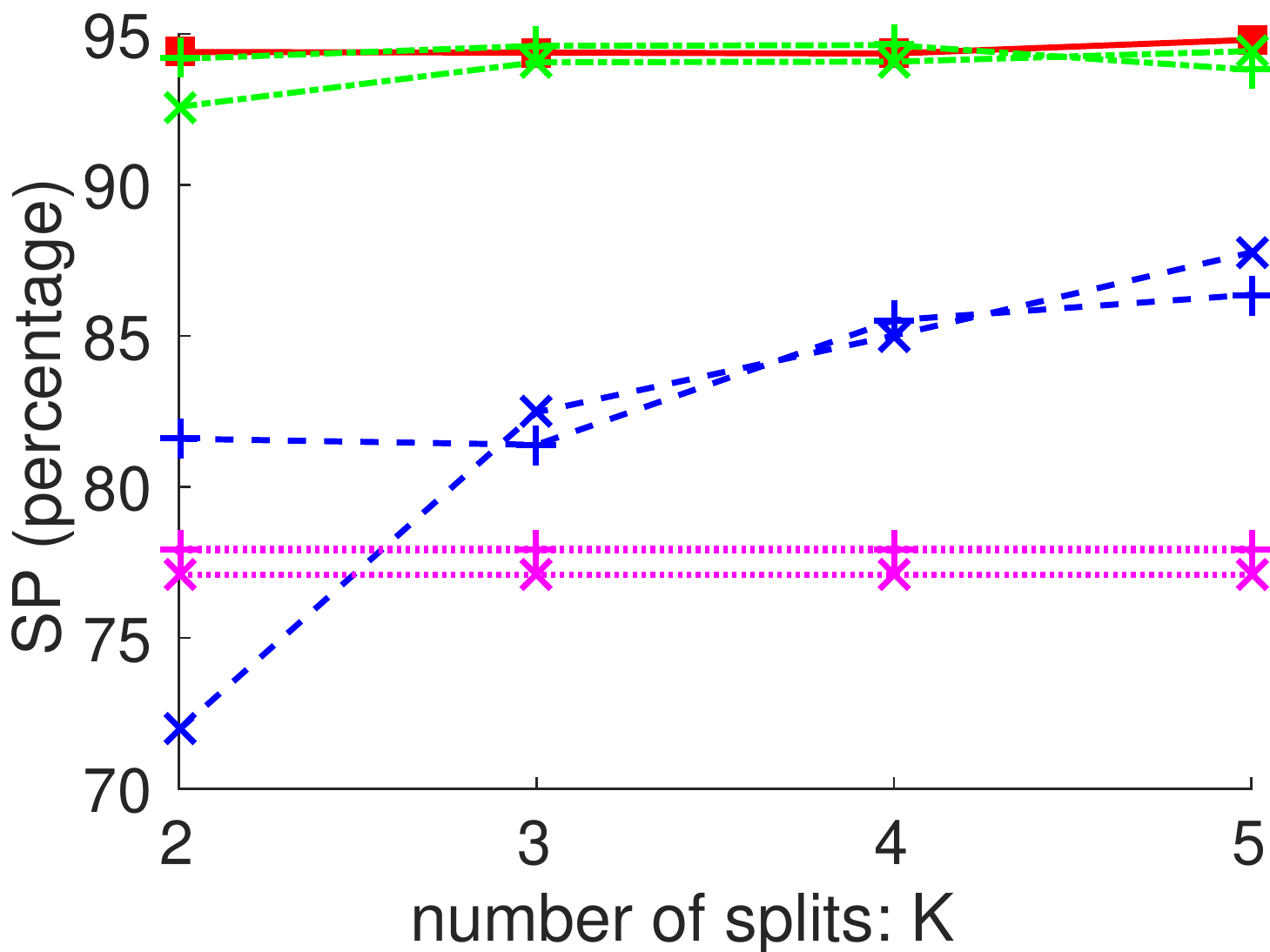} & 
\includegraphics[width=0.25\textwidth]{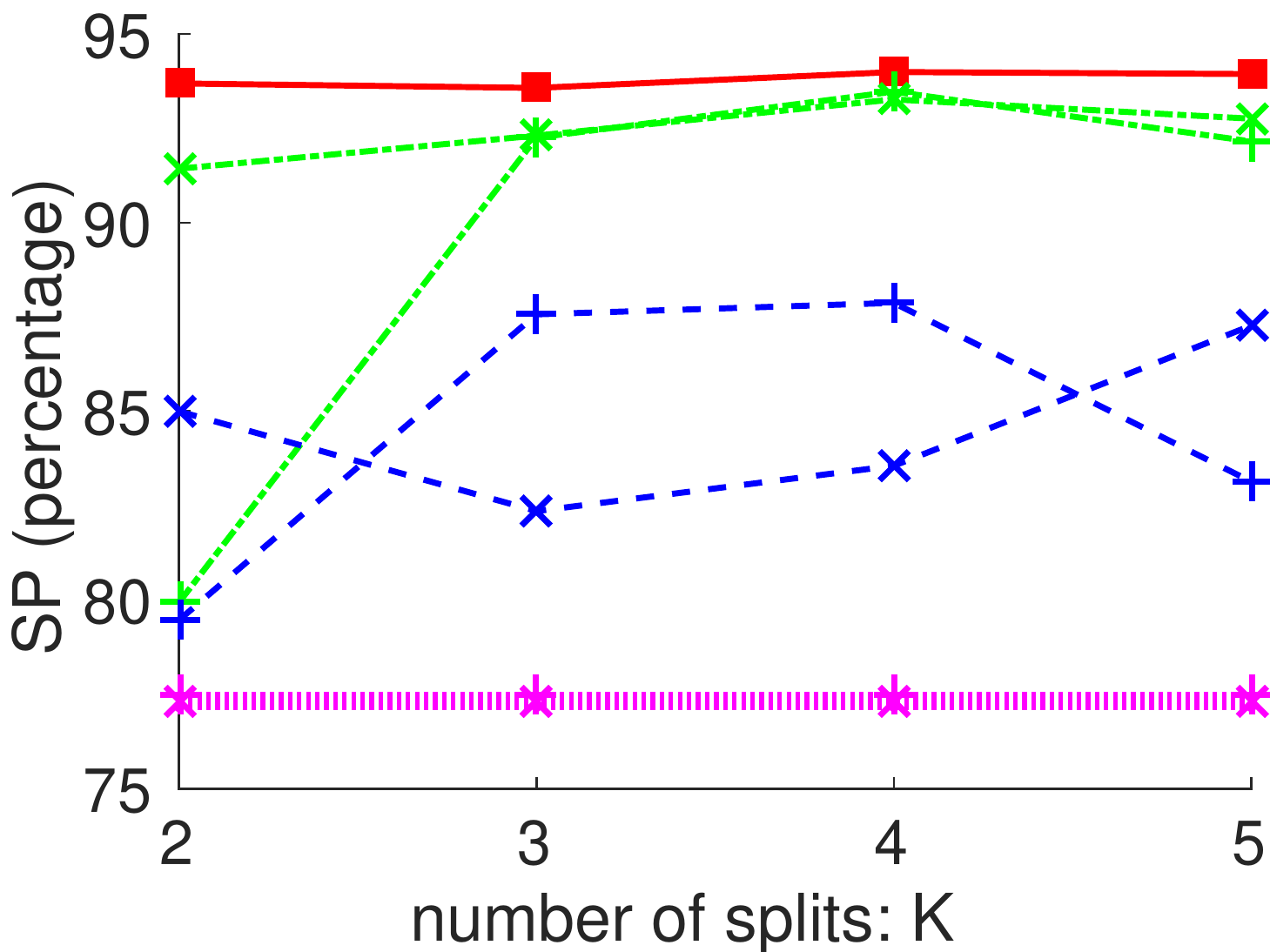} & 
\includegraphics[width=0.25\textwidth]{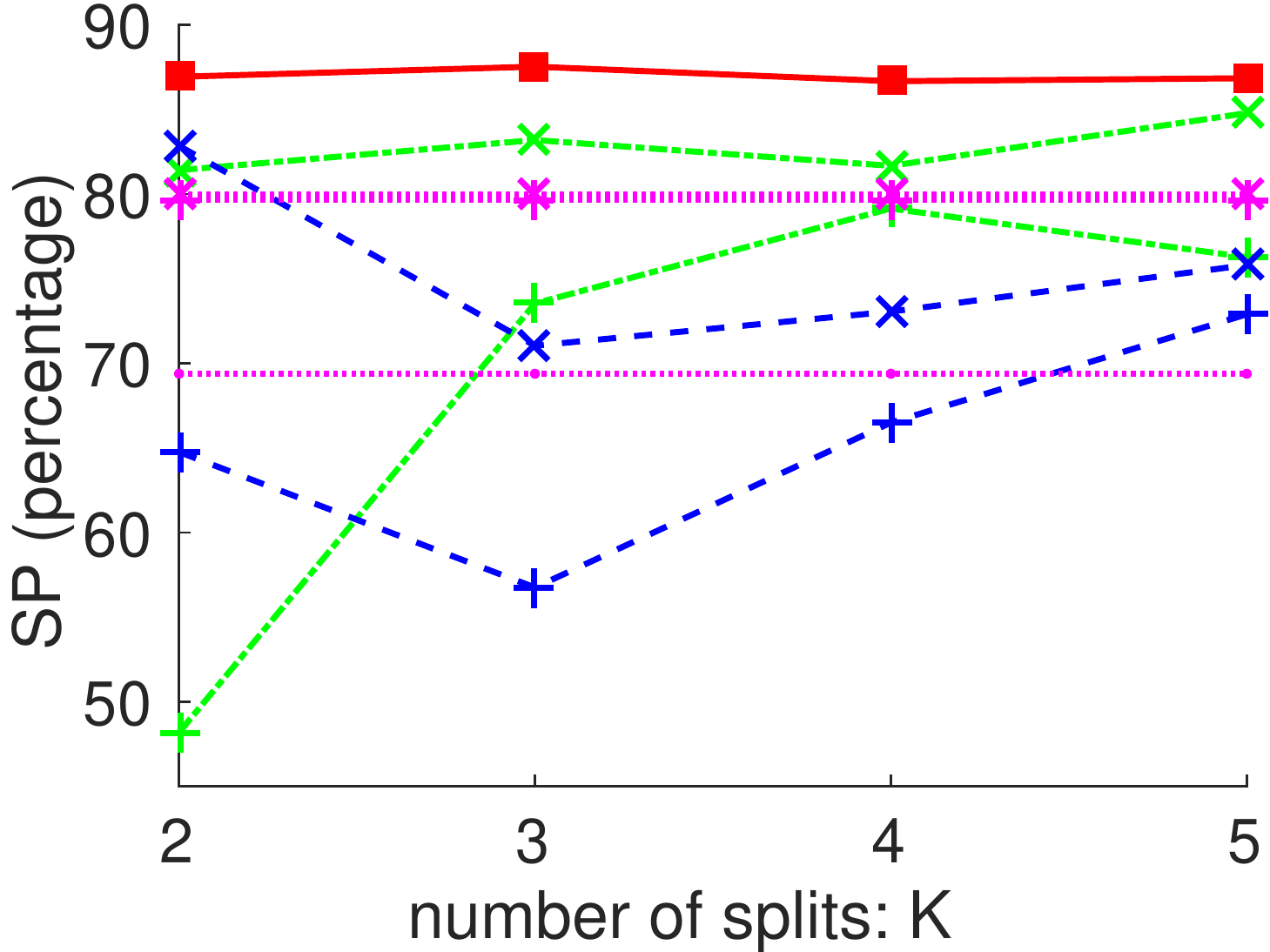} & 
\includegraphics[width=0.25\textwidth]{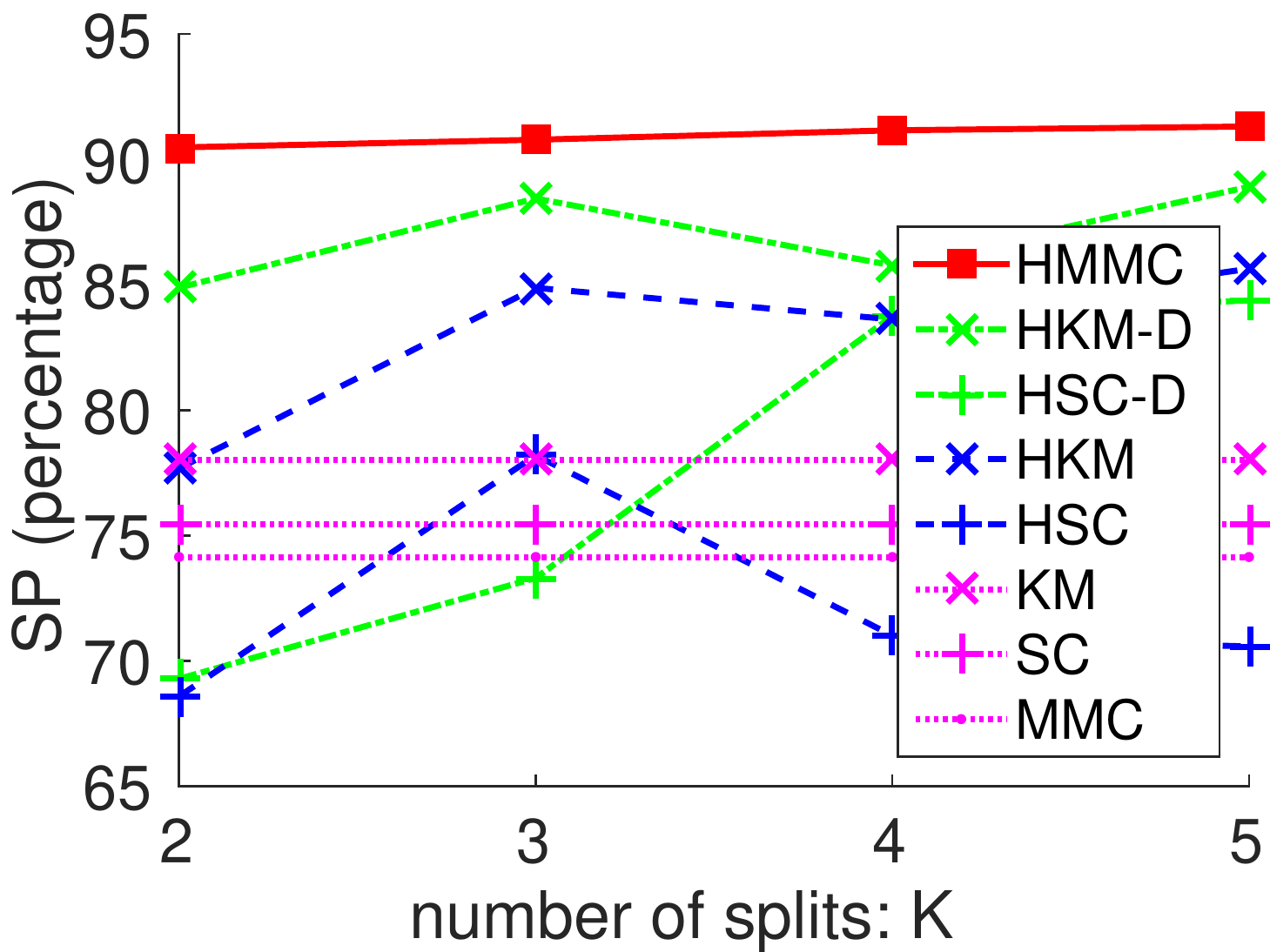} \\
(e) $K$ on AWA-ATTR & (f) $K$ on AWA-PCA & (g) $K$ on VEHICLE & (h) $K$ on IMAGENET
\end{tabular}
\caption{Using different settings of $F$ and $K$.  (a-d) plot against $F$ while fixing $K=2$, and (e-h) plot against $K$ while fixing $F$ as the number of ground-truth classes in each dataset.}
\label{fig:fk}
\end{figure*}

\textbf{Parameters}:  For a fair comparison of all the hierarchical top-down clustering methods, we apply the same stopping criterion: we test if the number of leaf nodes exceeds a fixed limit $F$.  Empirically, we set $F$ as 1, 1.5 and 2 times the number of ground-truth classes in each dataset.  The number of splits on each node also has a great impact on the learned hierarchy.  To compare different hierarchical clustering methods, we simply use $K$-nary branching for all splits in all hierarchies.  We experiment with $K$ set as 2, 3, 4 and 5, respectively.  With a particular setting of $F$ and $K$, we can fairly compare different hierarchical clustering methods since they perform the same number of splits and obtain the same number of leaf nodes.  We use the same solver for learning HMMC and its variants, and report the best performance with both $\alpha$ and $\beta$ selected from the range $\{10^{-4},10^{-3},10^{-2},10^{-1},10^{0}\}$.

For the HBUC baselines, we apply the same $F$ parameter as above.  However, all the HBUC methods use binary branching and there is no result for $K$ larger than 2.

For the flat clustering methods (\ie, KM, SC and MMC), we set the number of clusters to $F$ to fairly compare performance with hierarchical methods.  For SC, we use a 5-nearest neighborhood graph and set the width of the Gaussian similarity function as the average distance over all the 5-nearest neighbors.  This also applies in HSC and HSC-D which use SC for splitting a node.

\textbf{Performance Measures}:  We evaluate all the methods by three performance measures.  The first two are semantic measures focusing on how well the learned hierarchy captures the semantics in the ground-truth hierarchy.  The motivation is that two semantically similar images should be grouped in the same or nearby clusters in the learned hierarchy, and two semantically dissimilar images should be split into clusters that are far away from each other.

For each pair of images, we compute their semantic similarity from the ground-truth hierarchy, using the following two metrics.  The first \emph{shortest path} metric~\cite{Harispe13} finds the shortest path linking the two image classes in the ground-truth hierarchy, normalizes the path distance by the maximum distance, and subtracts the distance from 1 as the semantic similarity.  The second \emph{path sharing} metric~\cite{Fergus10} counts the number of nodes shared by the parent branches of the two image classes, normalized by the length of the longest of the two branches.  Note that we can similarly define the shortest path similarity and the path sharing similarity using the learned hierarchy, for any pair of images, by checking the leaf node(s) where the two images are clustered.  For flat clustering with no hierarchy, we simply set the similarity as 1 if two images are from the same cluster, and 0 otherwise.

To measure the goodness of the learned hierarchy in capturing semantics, we compute the mean squared error of the learned similarity and the ground-truth semantic similarity over all pairs of images, and subtract the mean squared error from 1 as our semantic measure.  Note that we have two semantic measures, the shortest path similarity (SP) and the path sharing similarity (PS).  The higher the values, the better the performance.

Moreover, we also report the Rand Index (RI)~\cite{Rand71}, which evaluates the percentage of true positives within clusters and true negatives between clusters.  Note that RI is a commonly-used measure for flat clustering.  For hierarchical clustering methods, we simply ignore the hierarchy and evaluate RI on the leaf node clustering (allowing direct comparisons with flat clustering methods).

\subsection{Results}
\label{sec:exper:result}

\begin{figure*}[t]
\footnotesize
\centering
\includegraphics[width=\textwidth]{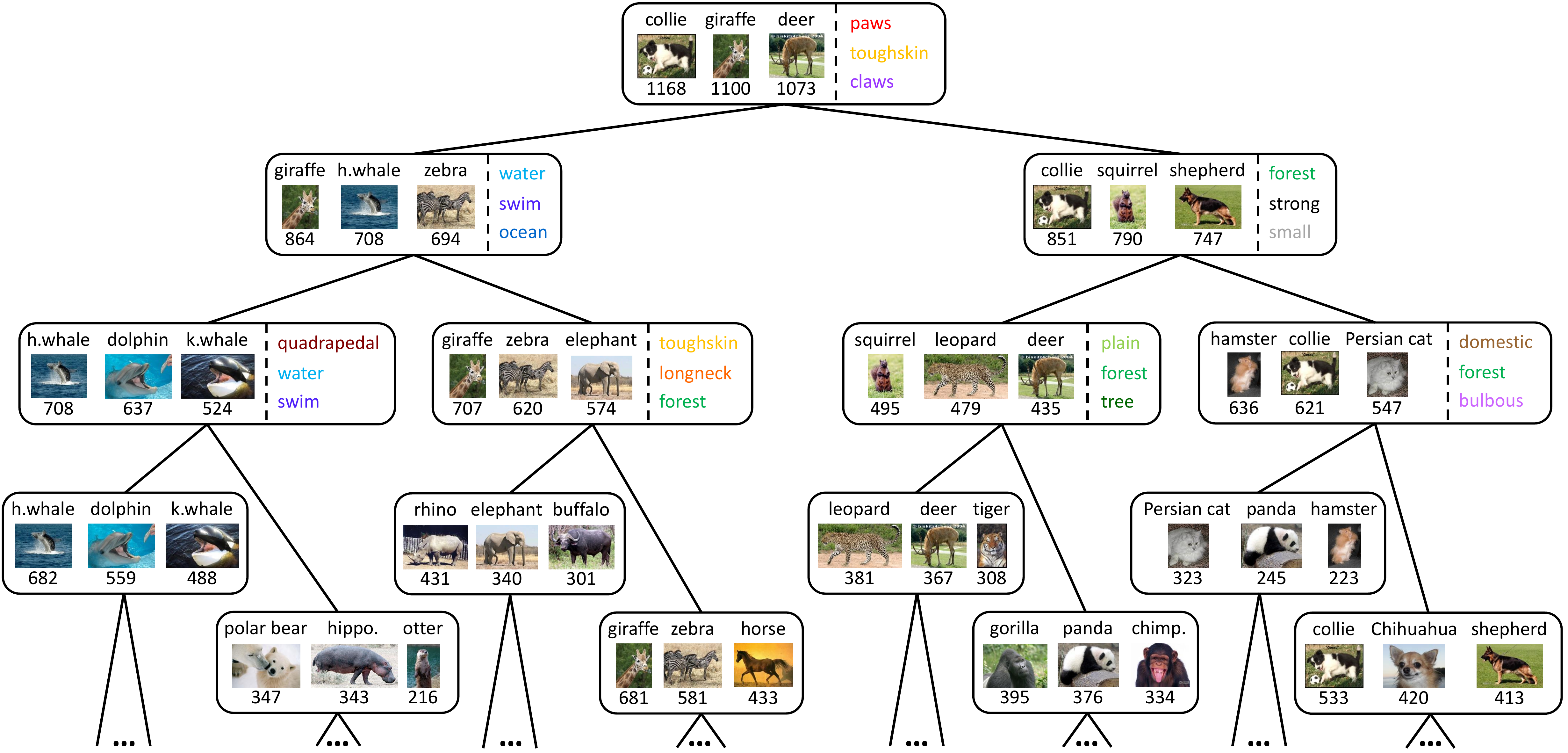}
\caption{(Best viewed in color.) The learned hierarchy on AWA-ATTR with binary branching.  Here we show the results on the first four layers.  For each node, we visualize three majority image classes (with the number of images from each class listed below the sample image), and the three most discriminative attributes (ranked by the magnitude of the regularization on the corresponding feature dimension).}
\label{fig:vis}
\end{figure*}

\textbf{Comparing flat and hierarchical methods}:  Due to space limitations, we only report the clustering results with $F$ equal to the number of ground-truth classes and $K=2$ (binary splitting).  The results are listed in Table~\ref{table:performance}, which shows that HMMC achieves the best performance on AWA-PCA, VEHICLE and IMAGENET, and competitive results on AWA-ATTR.  Specifically, HMMC improves over the second best by 0.2\% on AWA-ATTR, 2\% on AWA-PCA, 6\% on VEHICLE and 4\% on IMAGENET, respectively, in terms of the semantic measure SP.  This verifies that HMMC better captures the semantics in the clustered hierarchies.

Moreover, HMMC outperforms other HTDC baselines in most cases, showing the effectiveness of our greedy top-down algorithm for hierarchy building and our alternating descent algorithm for splitting data on a given node.  Note that the HBUC baselines tend to perform worse since they typically produced extremely unbalanced clusters at the top levels (\eg, a child contains only one sample).  They also did not lead to semantically meaningful hierarchies.


\textbf{Using different $F$ and $K$}:  We also vary the parameters $F$ (\ie, the number of leaf nodes) and $K$ (\ie, the number of splits), and plot the SP performance in Fig.~\ref{fig:fk}.  Here we have omitted the poor results of hierarchical bottom-up methods for better visualizations.  HMMC consistently outperforms the other baselines on AWA-PCA, VEHICLE and IMAGENET, and is comparable with HKM-D and HSC-D on AWA-ATTR.  Note that the performance of HMMC is stable with regard to the different settings of $F$ and $K$.

\begin{figure}[th]
\small
\centering
\includegraphics[width=0.45\textwidth]{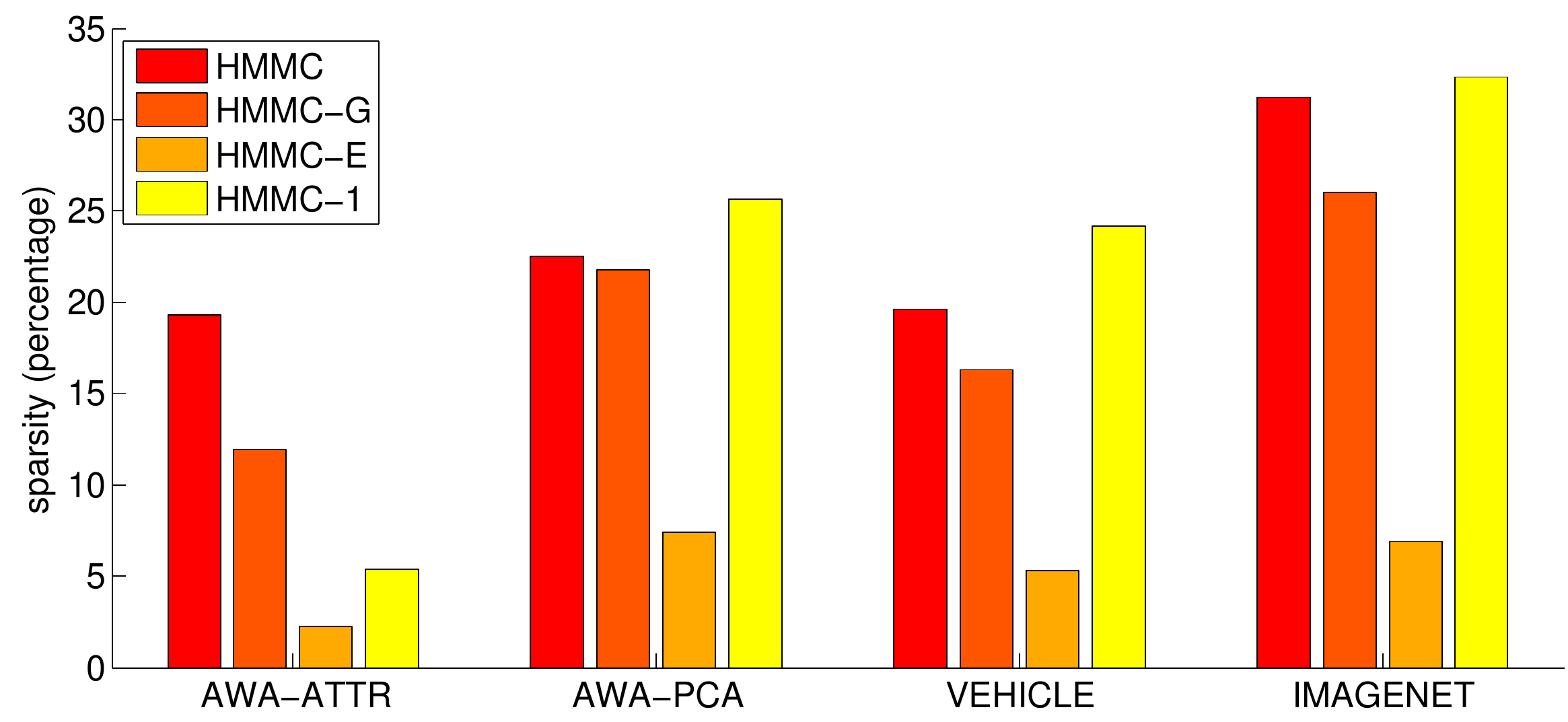} 
\caption{HMMC model sparsity.  See text for details.}
\label{fig:sparsity}
\end{figure}

\textbf{Comparing the variants of HMMC}:  Table~\ref{table:performance} shows that HMMC gets slightly better performance over the four variants of HMMC.  This is reasonable since HMMC produces sparse models that may better capture semantics.  We also compare the model sparsity (\ie, the percentage of zeros in the learned models) in Fig.~\ref{fig:sparsity}.  Here we omit HMMC-2 since the model is always non-sparse.  For a fair comparison, we fix the trade-off parameters to 1 in all models.  Note that by combining the grouping and exclusive regularizers, HMMC is sparser than HMMC-G and HMMC-E.  HMMC-1 sometimes has slightly better sparsity than HMMC, but the performance is limited due to the lack of semantics.

\textbf{Runtime comparison}: Table~\ref{table:performance} also reports the runtime results.  Our implementation of HMMC is between 1.4 to 8 times faster than MMC, showing the efficiency of the hierarchical method.  Note that HMMC is more expensive than other hierarchical and flat methods.  This is reasonable since HMMC needs to solve a more expensive optimization problem during clustering.

\textbf{Visualizations}: Fig.~\ref{fig:vis} visualizes the learned hierarchy on AWA-ATTR. See the caption for details.  Our model captures semantically meaningful attributes in building the hierarchy -- note how the attribute ``quadrapedal'' is used to separate ``whales'' and ``polar bears'', and how ``longneck'' is used to divide ``rhinos'' and ``giraffes''.

\section{Conclusion}
\label{concl}

We have presented a hierarchical clustering method for unsupervised construction of taxonomies.  We develop a greedy top-down splitting criterion, and use the grouping and exclusive regularizers for building semantically meaningful hierarchies from unsupervised data.  Our method makes use of maximum-margin learning, and we propose effective algorithms to solve the resultant non-convex objective.  We test our method on four standard datasets, showing the efficacy of our method in clustering, and the ability to capture semantics via the hierarchies.

\footnotesize{
\bibliography{ref}
\bibliographystyle{icml2015}
}
\end{document}